\documentclass[10pt,conference]{IEEEtran}
\IEEEoverridecommandlockouts
\usepackage{amsmath,amssymb,amsfonts}
\usepackage{algorithmic}
\usepackage{graphicx}
\usepackage{textcomp}
\usepackage{xcolor}
\usepackage{subcaption}

\def\BibTeX{{\rm B\kern-.05em{\sc i\kern-.025em b}\kern-.08em
    T\kern-.1667em\lower.7ex\hbox{E}\kern-.125emX}}

\begin{document}

\title{Handling big tabular data of ICT supply chains: a multi-task, machine-interpretable approach\\

\thanks{This work was supported in part by Mathematics of Information Technology and Complex Systems (MITACS) Accelerate Program, Smart Computing for Inovation Program (SOSCIP) and Lytica Inc.}
}

\author{\IEEEauthorblockN{Bin Xiao, Murat Simsek, Burak Kantarci}
\IEEEauthorblockA{\textit{School of Electrical Engineering and Computer Science} \\
\textit{University of Ottawa}\\
Ottawa, Canada \\
\{bxiao103, murat.simsek, burak.kantarci\}@uottawa.ca}
\and

\IEEEauthorblockN{Ala Abu Alkheir}
\IEEEauthorblockA{\textit{Directorate, Analytics} \\
\textit{Lytica Inc.}\\
Ottawa, Canada \\
ala\_abualkheir@lytica.com}
}

\maketitle

\begin{abstract}

Due to the characteristics of Information and Communications Technology (ICT) products, the critical information of ICT devices is often summarized in big tabular data shared across supply chains. Therefore, it is critical to automatically interpret tabular structures with the surging amount of electronic assets. To transform the tabular data in electronic documents into a machine-interpretable format and provide layout and semantic information for information extraction and interpretation, we define a Table Structure Recognition (TSR) task and a Table Cell Type Classification (CTC) task. We use a graph to represent complex table structures for the TSR task. Meanwhile, table cells are categorized into three groups based on their functional roles for the CTC task, namely Header, Attribute, and Data. Subsequently, we propose a multi-task model to solve the defined two tasks simultaneously by using the text modal and image modal features. Our experimental results show that our proposed method can outperform state-of-the-art methods on ICDAR2013 and UNLV datasets.  
\end{abstract}

\begin{IEEEkeywords}
Big Data Analytics, Supply Chain Optimization, Image Processing, Table Structure Recognition, Table Cell Type Classification
\end{IEEEkeywords}

\section{Introduction}
\label{sec:introduction}

\begin{figure}
     \centering
     \begin{subfigure}[b]{\columnwidth}
         \centering
         \includegraphics[width=\columnwidth]{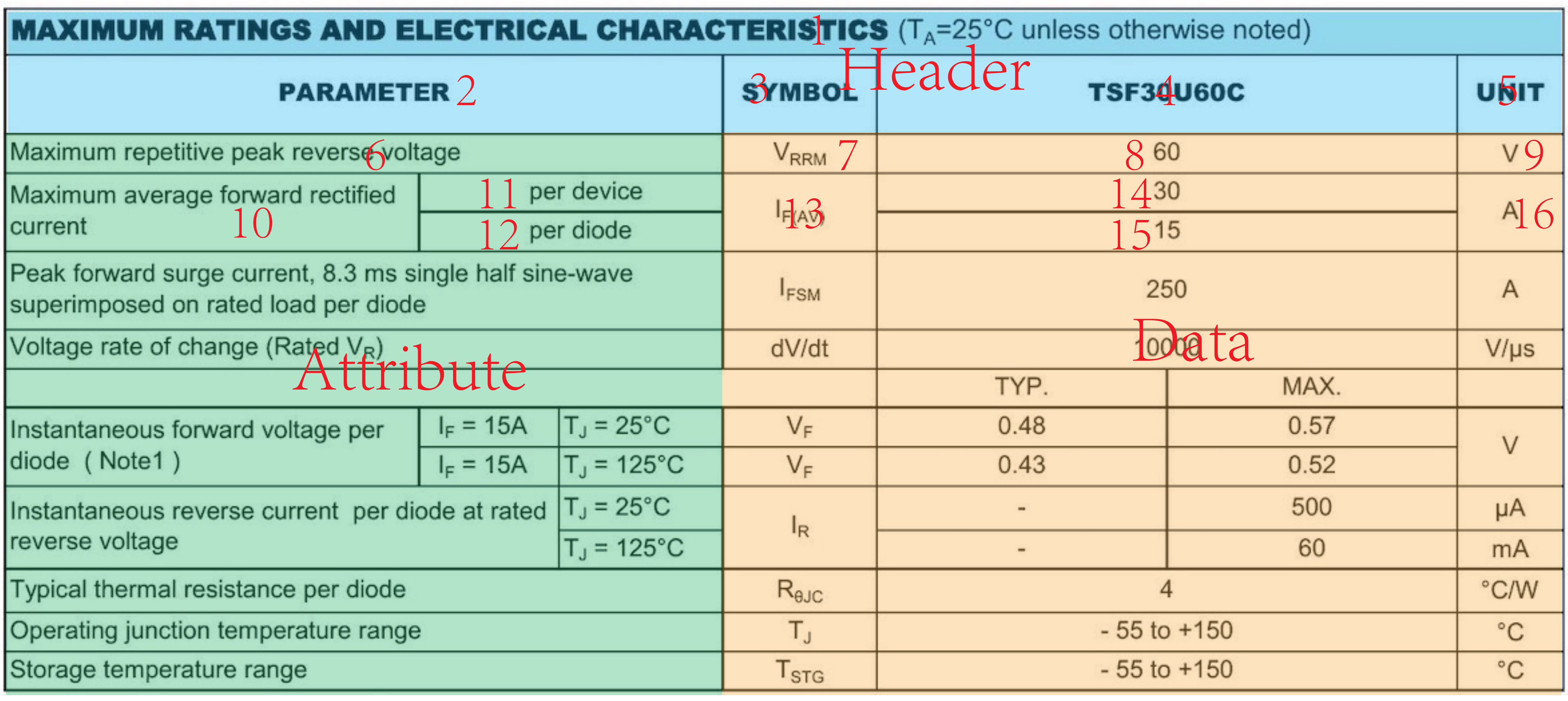}
         \caption{}
         \label{fig:ctc_sample}
     \end{subfigure}
     \hfill
     \begin{subfigure}[b]{0.5\columnwidth}
         \centering
         \includegraphics[width=\columnwidth]{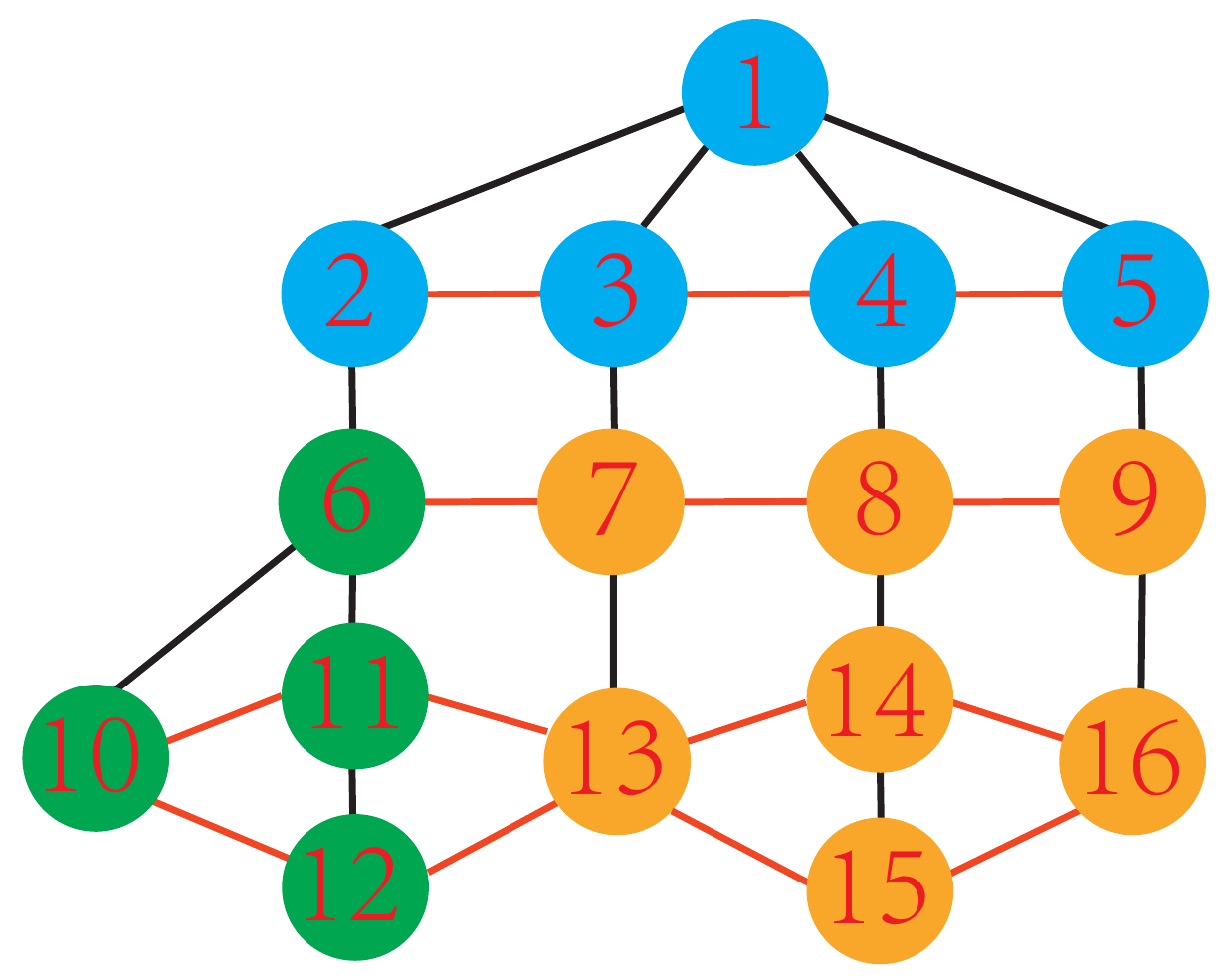}
         \caption{}
         \label{fig:tsr_sample}
     \end{subfigure}
    \caption{Figure (a) shows three types of table cells, namely header, attribute, and data of an ICT device. Figure (b) is part of the table cells' graph representation, in which each vertex represents a cell, red lines stand for horizontal connection, and black lines stand for vertical connection.}
    \label{fig:cell_type_sample}
\end{figure}

In Information and Communications Technology (ICT) supply chains, electronic devices often contain various parameters, units, or other critical information formatted in tables, making it vital to extract and interpret tables from electronic documents. Even though tables in electronic documents are user-friendly for human readers, they are often not structured and not machine-interpretable. It is also not practical for humans to read, extract and interpret tables from millions of electronic documents. Therefore, to deal with the vast amount of electronic documents in the ICT supply chain and make unstructured tabular data machine-interpretable, we define a Table Structure Recognition (TSR) problem, which can recover a complex table structure with a graph, and a Table Cell Type Classification (CTC) problem based on the cell's functional roles. More specifically, for the TSR problem, each cell in a table is represented with a vertex in a graph, and three types of cell associations, namely vertical connection, horizontal connection, and no connection, are defined to represent the locational relations among table cells, which can be represented by the edges in a graph. Meanwhile, similar to some existing studies~\cite{koci2016cell,gol2019tabular,wang2021tuta} discussing tabular cell classification problems with different taxonomies of cell types, we define three types of cells, namely Header, Attribute, and Data. Figure~\ref{fig:cell_type_sample} shows a sample table with defined types in the CTC problem and its graph representation of the table numbered part. Typically, headers in a table can express the meaning of their corresponding columns, attributes represent the meaning of their corresponding rows, and data cells are used to present the exact information of the specific header and attribute. In other words, the facts and information contained in a table can become accessible and easily located and extracted based on the defined three types of cells and the machine-readable table structure. It is worth mentioning that most of the existing studies~\cite{koci2016cell,gol2019tabular,wang2021tuta} on CTC problems focus on tables in spreadsheets which is a much easier problem definition because spreadsheets can provide more meta-information and the default units in the spreadsheets are cells. There are some studies~\cite{lohani2018invoice, tang2021matchvie} trying to extract entities and information from images, which is similar with our defined CTC problem, but they are not focusing on tables in document images, meaning that their problem definitions are not suitable to be solved together with TSR problem in a multi-task manner.

The main contribution of this study is three-fold: 1) This study extends the CTC problem to the tables in Portable Document Format (PDF) documents and image documents and builds a benchmark for this problem. 2) We propose a multi-task approach that simultaneously solves the defined TSR and CTC problems, achieving state-of-the-art results. Experimental results show that our proposed method can increase the F1 score from 70.76\% to 88.17\%, from 71.90 to 83.74 for the CTC task on the ICDAR2013 and UNLV datasets, respectively. For the TSR task, the proposed method can increase the F1 score from 92.30\% to 93.04\% on the ICDAR2013 dataset and from 87.24\% to 89.51\% on the UNLV dataset. 3) We discuss the different aspects of the proposed method and show their effectiveness with experiments. 

This paper is organized as follows: Section~\ref{sec:related_work} discusses the related studies. Section~\ref{sec:proposed_method} presents the full process of transforming the tabular data into the machine-interpretable format and presents the proposed multi-task method. Section~\ref{sec:experiments} presents and discusses the experimental results. We draw our conclusion and possible directions in section~\ref{sec:conclusion}.

\section{Related work}
\label{sec:related_work}

To extract and interpret critical information from the surging amount of electronic documents in the global ICT supply chain, it is vital but often challenging to transform unstructured tabular data into structured and machine-interpretable format, because it needs several separate steps to finish this task. Since the dominant studies in recent years first transform PDF files into document images and then use deep learning based methods, we only focus on deep learning based approaches.

\subsection{Tabular Structure Recognition}
TSR aims to identify table cells in identical rows and columns, which is challenging because of the complex table structures, such as merging cells and non-explicit border lines. Using a graph to represent the complex table structure is a popular design, in which table cells are represented by the graph nodes, and the associations of cells are defined in three types: vertical connection, horizontal connection and no connection, and graph edges are used to represent the defined cell associations. Many studies~\cite{Adiga2019TableSR,Chi2019ComplicatedTS} follow this problem formulation and propose bottom-up approaches. In contrast, some work, such as DeepDeSRT~\cite{Schreiber2017DeepDeSRTDL}, CascadeTabNet~\cite{tensmeyer2019deep} and TableDet~\cite{fernandes2021tabledet}, using top-down approaches would define the structural recognition as object detection or segmentation problem, often together with table detection problem. Similarly, these methods often employ and extend Cascade R-CNN~\cite{Cai2018CascadeRD}, Mask-RNN~\cite{he2017mask} and utilize transfer learning and data augmentation methods to improve the performance. 

\subsection{Table Cell Type Classification}
Many studies employ text embeddings and stylistic features for the spreadsheet's CTC problem. Elvis et al., in their work~\cite{koci2016cell} use Weka~\cite{hall2009weka} to select features and try various tree-based classifiers. Majid et al., in their work~\cite{gol2019tabular}, propose to incorporate pre-trained cell embedding and stylistic features. They design a neighbor-based approach to generate cell contexts, propose an embedding model built on the InferSent model~\cite{infersent}, and build an LSTM based classifier. Language models also have been used to the tabular cell classification problem. Training a large language model requires a large corpus and many computation resources. In order to apply a language model to the CTC task, the pre-trained language model is firstly used to generate the feature embeddings and fine-tune a cell type classification model. TUTA~\cite{wang2021tuta} is a large language model for a generally structured table trained with a large spreadsheets corpus. Besides, other language models, such as BERT~\cite{Devlin2019BERTPO} also can be used for the CTC problem. All these studies focus on Spreadsheets, which have inherent structural information that can help to train and fine-tune context-based language models. Conversely, tabular data in PDF documents do not contain this type of internal structural information, making it a more challenging task.

\section{Proposed Method}
\label{sec:proposed_method}
\begin{figure}[htp]
\begin{center}
  \includegraphics[width=\columnwidth]{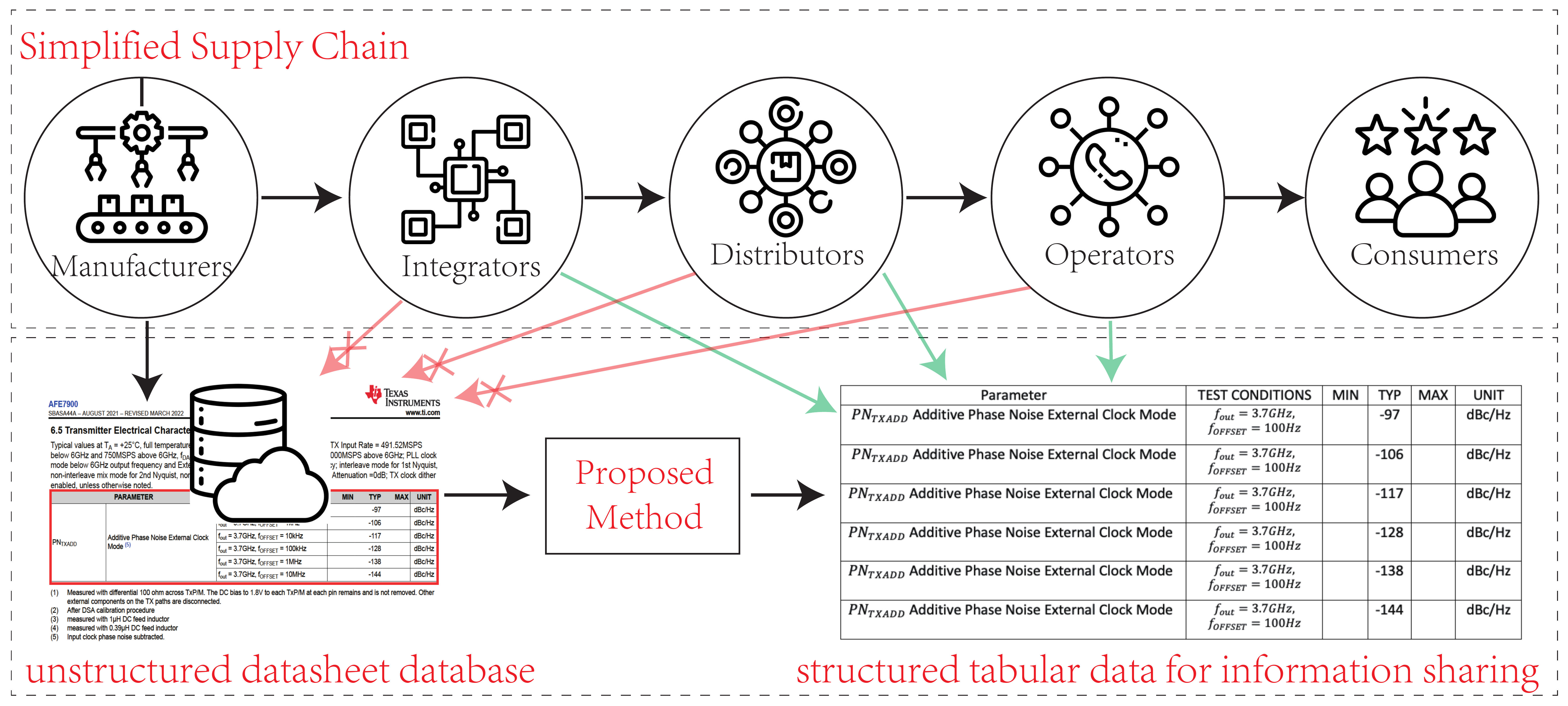}
  \caption{A simplified ICT supply chain sample.}
  \label{fig:supply_chain_sample}
\end{center}
\end{figure}

As illustrated in Fig. \ref{fig:supply_chain_sample}, manufacturers generate "big tabular data" with key information in tables but are hard to be shared with other participators in the supply chain. Our proposed method transforms unstructured tabular information into structured and machine-interpretable format, paving the way to efficient information sharing in the whole supply chain.

\subsection{Problem Definition}
\label{sec:problem_definition}
TSR aims at recovering complex table structures into a structured and machine-interpretable format. Following the problem definition in some existing bottom-up approaches ~\cite{Chi2019ComplicatedTS}, we also use a graph to represent a complex table structure, in which graph vertexes are used to represent table cells and graph edges are use to denote the associations between two cells in a table, as shown in Figure \ref{fig:cell_type_sample}. Assuming that each table cell's bounding box is given, which can be denoted by $\{x_i^1, x_i^2, y_i^1, y_i^2\}$, where $i$ is the cell number. Then, the $k$th table $t_k$ in a table set $\mathbf{T} = \{t_k; k \in K \}$ containing $K$ tables can be represented by a set of cells $\mathbf{C_k}=\{c_{i}^k; i \in N_k\}$, where $c_i^k$ stands for the $ith$ cell in the table $t_k$, $N_k$ is the number of cells in the table $t_k$. Thus, for the table structure recognition problem, given a training set consists of $K$ tables , and the table $t_k$'s cell association set $\mathbf{R_k}=\{ r_{\{i,j\}}^k; i \neq j, i \in N_k, j \in N_k, k \in K \}$, where $r_{\{i, j\}}^k=\{\{c_i^k, c_j^k\}; i \neq j, i \in N_k, j \in N_k, k \in K\}$, and the corresponding label set $Y_k^{tsr}=\{y_{\{i, j\}}^k; y \in \{0, 1, 2\}, i \in N_k, j \in N_k, k \in K\}$ of the association set, where $0, 1, 2$ means the two cells in the association are horizontal connected, vertical connected and not connected. Therefore, the task of table structure recognition is to train a predictive model with given training data that can determine the probability of the two cells relation in a cell association, namely $P_\theta(y_{\{i,j\}}^k =\hat{y}_{\{i,j\}}^k \rvert r_{\{i,j\}}^k), \hat{y}_{\{i,j\}}^k \in\{0, 1, 2\}$.

\begin{table}[ht!]
\caption{A summary of notations used in this paper.}
\centering
\begin{tabular}{ c c}
\hline
\label{table:notation_table}
 $\mathbf{T}$ & A table set \\
 $\mathbf{C_k}$ & The cell set of the $k$th table\\
 $\mathbf{R_k}$ & The cell association set of the $k$th table\\
 $\mathbf{Y_k^{tsr}}$ & The label set of the $k$th table for the TSR\\
 $\mathbf{Y_k^{ctc}}$ & The label set of the $k$th table for the CTC\\
 $\{x_i^1, x_i^2, y_i^1, y_i^2\}$ & The coordinate of the $i$th cell in a table\\
 $c_i$ & The $i$th cell in a table\\
 $t_k$ & The $k$th table in a table set $\mathbf{T}$ containing $K$ tables\\
 $r_{\{i,j\}}^k$ & The association of the $i$th, $j$th cell in the $k$th table\\

 $\mathcal{L}$ & The loss function \\
 $\theta, \phi, \omega, \sigma$ & The trainable parameters of each function\\
 $\mathcal{F}$ & A fully connected layer\\
 $\mathbf{f}$ & The output of a fully connected layer\\
 $\mathcal{CAT}$ & The function that can represent the fusion layer\\
 $\mathcal{C}$ & The classifier\\
 $\mathcal{E}$ & The embedding network\\
 $\mathbf{x}$ & The input of a function\\
 $\mathbf{y}$ & The output of the proposed method\\
 $\mathbb{R}^{ch*h*w}$ & A real value set with $ch$ channel, $h$ height, $w$ width\\
 \hline
\end{tabular}
\end{table}

CTC problem aims at identifying their functional roles in a complex table structure. For the table cell classification program, given a training set consists of $K$ tables, each table $t_k$ has a set of cells $\mathbf{C_k}=\{c_i^k; i \in N_k\}$ and their corresponding types $\mathbf{Y_k^{ctc}}=\{y_i^k; i \in N_k \}$. We define three types of cells, namely Header, Attribute, Data, meaning that $y_i^k \in \{0, 1, 2\}$. Therefore, the task of table cell classification is to train a predictive model with given training data that can determine the probability of a cell belonging a predefined cell types, namely $P_\phi(y_i^k =\hat{y}_i^k \rvert c_i^k\}, \hat{y}_i^k \in\{0, 1, 2\}$. Notably, the proposed method is a bottom-up approach and requires that each cell's bounding boxes are known. In our implementation, we use MMOCR~\cite{kuang2021mmocr} to detect table cells and extract the content from table images. The notations used in this paper are summarized in Table~\ref{table:notation_table}.

\subsection{Multi-task Table Structure Recognition and Table Cell Type Classification}
\label{sec:multi_task_tsr_ctc}

\begin{figure*}[htp]
\begin{center}
  \includegraphics[width=0.95\textwidth]{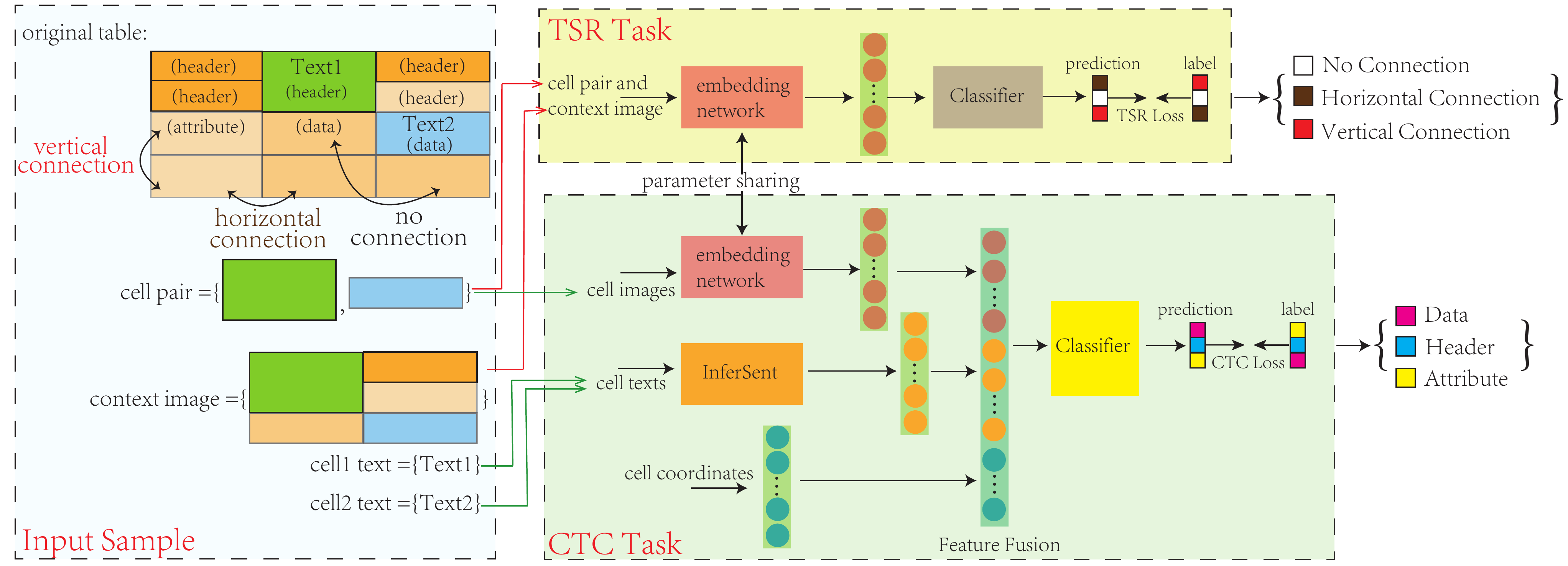}
  \caption{Overall architecture of the proposed multi-task model. Notably, header, attribute and data are three types of cell in the CTC task, and horizontal connection, vertical connection and no connection are three possible outputs in the TSR task. }
  \label{fig:tsr_ctc_overall_architecture}
\end{center}
\end{figure*}
To transform unstructured tabular data into a structured and machine-interpretable format, we need three steps: table detection, table structure recognition, and table cell type classification. Luckily, there have been some studies that achieved promising results on the table detection, as discussed in section~\ref{sec:related_work}. In practice, we first convert all the PDF documents into document images and then use TableDet~\cite{fernandes2021tabledet} to extract all the tables from the document images such that we can obtain the table set $T=\{t_k; k \in K\}$. To generate the cell association set required by the table structure recognition task, we follow the popular KD-tree based K-nearest method, which is also used in existing studies~\cite{Chi2019ComplicatedTS,xiao2022table}. More specifically, we can calculate the distance between two cells by a pre-defined distance metric, usually Euclidean Distance, using their bounding boxes. Thus, for the $i$th cell in the $k$th table $t_k$, we can find its nearest top $M$ neighbor cells to form association pairs, and a KD-Tree model can easily implement this process. Notably, $M$ is a hyperparameter set to 20 in our implementation. 

Even though PDF documents and document images can only provide very limited meta-information, they can inherently provide visual and text information, meaning that features can be extracted from both text and image modalities. Following the typical design of multi-task approaches, there are two branches in the proposed model, one is for the TSR task, and another is for the CTC task, as shown in Figure~\ref{fig:tsr_ctc_overall_architecture}. The TSR branch only utilizes the features from the image modal, which can be extracted by an embedding network. In contrast, the CTC branch uses features from both image modal, text modal, and cells' coordinates.
More specifically, in the TSR branch, we follow the design of study~\cite{xiao2022table}, crop both cells' images in each cell association pair and their context image, and then use the cropped cell images and the context image as the input in the table structure recognition branch, as shown in the input sample part of Figure~\ref{fig:tsr_ctc_overall_architecture}. Then the generated images are fed into an embedding network, denoted by the function $\mathcal{E}_\phi$, to extract visual feature maps. Then the extracted feature maps are fed into a fully connected layer which can be represented by a function $\mathcal{F}_{\theta_1}^{tsr}$. At last, the outputs of $\mathcal{F}_{\theta_1}^{tsr}$ are further fed into the classifier $\mathcal{C}_{\sigma_1}^{tsr}$ with a softmax function to get the probability of each class. The full process can be represented by the Equation~\ref{eq:tsr_process}, where $ch, h, w$ are the number of channels, image height, and image width, respectively, and all of $\theta, \sigma, \phi$ are trainable parameters.

\begin{equation}
\label{eq:tsr_process}
\begin{aligned}
\mathbf{y}_{tsr} = \mathcal{C}_{\sigma_1}^{tsr}(\mathbf{f}_{tsr}), \mathbf{f}_{tsr} = \mathcal{F}_{\theta_2}^{tsr}(\mathcal{E}_\phi(\mathbf{x}^{tsr})), \\ \mathbf{x}_{tsr} \in \mathbb{R}^{ch*h*w}, \mathbf{f}_{tsr} \in \mathbb{R}^{l_1}, \mathbf{y}_{tsr} \in \mathbb{R}^3
\end{aligned}
\end{equation}

Meanwhile, in the CTC branch, together with the cells' coordinates, the cropped cell images and their corresponding texts are used as the inputs of the CTC branch. Assuming that the InferSent~\cite{infersent} module can be represented by function $\mathcal{I}_\omega$, the fully connected layer for the text modal can be denoted by function $\mathcal{F}_{\theta_2}^{ctc}$, then the text features can be calculated by the Equation~\ref{eq:ctc_text_process}, where $\mathbf{f}_{2}$ is the output feature and $l_{text}$ is the number of tokens in the input sentence.

\begin{equation}
\label{eq:ctc_text_process}
\begin{aligned}
\mathbf{f}_{text} = \mathcal{F}_{\theta_2}^{ctc}(\mathcal{I}_\omega(\mathbf{x}_{text})), \mathbf{x}_{text} \in \mathbb{R}^{l_{text}}, \mathbf{f}_{text} \in \mathbb{R}^{l_2}
\end{aligned}
\end{equation}

Similarly, we can obtain the image features and the coordinate features in the CTC branch using the Equation~\ref{eq:ctc_image_process} and~\ref{eq:ctc_coordinate_process}. At last, all features are fused together with fusion function $\mathcal{CAT}$ and fed into the classifier $\mathcal{C}_{\sigma_2}^{ctc}$, which can be achieved by Equation~\ref{eq:ctc_process}. It is worth mentioning that $l_1, l_2, l_3$ and $l_4$ are determined by the number of neurons in each fully connected layer, and $\mathcal{E}_\phi$ shares parameters in both branches.

\begin{equation}
\label{eq:ctc_image_process}
\begin{aligned}
\mathbf{f}_{img} = \mathcal{F}_{\theta_2}^{ctc}(\mathcal{E}_\phi(\mathbf{x}_{img})), \mathbf{x}_{img} \in \mathbb{R}^{c*h*w}, \mathbf{f}_{img} \in \mathbb{R}^{l_3}
\end{aligned}
\end{equation}

\begin{equation}
\label{eq:ctc_coordinate_process}
\begin{aligned}
\mathbf{f}_{coord} = \mathcal{F}_{\theta_3}^{ctc}(\mathbf{x}_{coord}),
\mathbf{x}_{coord} \in \mathbb{R}^{4}, \mathbf{f}_{coord} \in \mathbb{R}^{l_4}
\end{aligned}
\end{equation}

\begin{equation}
\label{eq:ctc_process}
\begin{aligned}
\mathbf{y}_{ctc} = \mathcal{C}_{\sigma_2}^{ctc}(\mathcal{CAT}(\mathbf{f}_{img}, \mathbf{f}_{text}, \mathbf{f}_{coord})),  \mathbf{y}_{ctc} \in \mathbb{R}^3
\end{aligned}
\end{equation}

Since the proposed method employs multi-task architecture, meaning that all the parameters should be trained together, the loss function $\mathcal{L}_{mt}$ can be defined as Equation~\ref{eq:loss_function}, which contains a TSR loss denoted by $\mathcal{L}_{tsr}$, a CTC loss denoted by $L_{ctc}$ and a hyper parameter $\lambda$ to balance $\mathcal{L}_{tsr}$ and $\mathcal{L}_{ctc}$. We use cross-entropy loss for both $\mathcal{L}_{tsr}$ and $\mathcal{L}_{ctc}$ in our implementation, which can be defined as Equation~\ref{eq:cross_entropy}, where $y_i$ means the $i$th prediction and $\hat{y}_i$ is its corresponding ground truth.

\begin{equation}
\label{eq:loss_function}
\begin{aligned}
\mathcal{L}_{mt} = (1 - \lambda) * \mathcal{L}_{tsr} + \lambda * \mathcal{L}_{ctc} 
\end{aligned}
\end{equation}

\begin{equation}
\label{eq:cross_entropy}
\begin{aligned}
\mathcal{L}_{ce}(\mathbf{y}, \mathbf{\hat{y}}) = -\sum_{i=1}^N \hat{y}_i \log y_i
\end{aligned}
\end{equation}

\section{Experiments and Analysis}
\label{sec:experiments}

ICDAR2013 and UNLV datasets are popular for table detection and TSR problems, containing 156 and 540 tables, respectively. We further annotate these two datasets with "Header," "Attribute," or "Data," as discussed in section~\ref{sec:problem_definition}, and exclude all the empty cells in the CTC task labeling because empty cells can be easily identified without using any machine learning models. In the experiments, the ICDAR2013 dataset is randomly split into a validation set with 33 tables and a testing set with 123 tables. The UNLV dataset is also randomly split into a training, validation, and testing set with 323, 107, 110 tables, respectively. Notably, we remove 18 tables from the UNLV dataset because of the ambiguity of their labels, and we use the training set of the UNLV dataset when evaluating the performance on the ICDAR2013 dataset, following the experimental setting of study~\cite{xiao2022table}. We use MMOCR~\cite{kuang2021mmocr} to extract text contents from the table cells for both ICDAR2013 and UNLV datasets.

\subsection{Implementation Details and Experimental Results}
\label{sec:expeirmental_results}
As discussed in section~\ref{sec:multi_task_tsr_ctc}, the proposed method contains an embedding network $\mathcal{E}_\phi$, which is used in both the TSR task and CTC task. In our implementation, we use a simple ConvNet-4~\cite{vinyals2016matching}, which consists of four convolution layers. Following the method described in the study~\cite{xiao2022table}, the input images of the embedding network are firstly resized to the dimension $3*84*84$ with zero padding to keep the original height-width ratio. Then the resized images are fed into the embedding network $\mathcal{E}_\phi$ followed by a fully connected layer $\mathcal{F}$ to obtain the features. In our implementation, we set the number of neurons in both $\mathcal{F}_{\theta_1}^{tsr}$ and $\mathcal{F}_{\theta_2}^{ctc}$ to 128, meaning that both ${l_1}$ and ${l_3}$ equal 128. Similarly, $l2$ and $l4$ are set to 128 and 32, respectively, and $\lambda$ in the loss function is set to 0.3 for the ICDAR2013 dataset and 0.6 for the UNLV dataset. Each classifier, as shown in Figure~\ref{fig:tsr_ctc_overall_architecture}, is implemented by a fully connected layer with three neurons followed by a softmax function to transform the logits into a distribution. We use a simple concatenation function to implement the fusion function $\mathcal{CAT}$ which does not have any trainable parameters. Notably, since the datasets used in the experiments are unbalanced, we use a cost-sensitive method in the implementation of the loss functions.

For the TSR task, we list TabbyPDF~\cite{shigarov2018tabbypdf}, GraphTSR~\cite{Chi2019ComplicatedTS}, DeepDeSRT~\cite{Schreiber2017DeepDeSRTDL} and CATT-Net~\cite{xiao2022table} as benchmark models. Considering that our proposed multi-task model relies on features from both image modal and text modal, we follow the popular pre-trained fine-tune paradigm, implement four models by fine-tuning the Glove~\cite{pennington2014glove}, FastText~\cite{mikolov2018advances}, BERT~\cite{Devlin2019BERTPO} to compare the state-of-the-art methods in the NLP. In terms of image modal, we implement two image classification models, namely ConvNeXt~\cite{liu2022convnet} and ResNet50~\cite{he2016deep} without using pre-trained weights. BERT is a popular context-based language model, meaning that the performance of these two models heavily relies on long dependency in a context, whereas Glove and FastText are two typical non-context-based models. ResNet50 is a classic convolution network that is widely used in image classification problems, while ConvNeXt~\cite{liu2022convnet} is the state-of-the-art model for the image classification problem. Since we only focus on the tables and use TableDet~\cite{fernandes2021tabledet} to extract all the tables from the document images without considering the context information of the tables, fine-tuning BERT shows very poor performance for the CTC problem; hence not included in Table~\ref{table:ctc_results}.

Table~\ref{table:tsr_results} and Table~\ref{table:ctc_results} show the overall Precision, Recall, and F1-score of the proposed method for the TSR task and CTC task, respectively. The performance scores of TabbyPDF, GraphTSR, and DeepDeSRT on the ICDAR2013 dataset come from the study~\cite{Chi2019ComplicatedTS}. "-" in Table~\ref{table:tsr_results} means the score is not reported in related studies. The experimental results show that the proposed method can outperform the benchmark models in the TSR and CTC tasks.
\begin{table}[ht!]
\caption{Experimental results for the TSR task.}
\centering
\begin{tabular}{  c | c | c | c || c | c |c }
\hline
\label{table:tsr_results}
& \multicolumn{3}{c}{ICDAR2013} & \multicolumn{3}{c}{UNLV}\\ 
\hline
Method & Prec & Recall & F1 & Prec & Recall & F1\\
\hline
 \texttt{TabbyPDF} & $ 78.90 $ & $ 84.50 $ & $ 81.60 $ & $ - $ & $ - $ & $ - $ \\
 \texttt{GraphTSR} & $ 81.90 $ & $ 85.50 $ & $ 83.70 $ & $ - $ & $ - $ & $ - $ \\
 \texttt{DeepDeSRT} & $ 57.30 $ & $ 56.40 $ & $ 56.80 $ & $ - $ & $ - $ & $ - $ \\
 \texttt{CATT-Net} & $ \mathbf{94.10} $ & $ 90.70 $ & $ 92.30 $ & $ 86.28 $ & $ \mathbf{88.31} $ & $ 87.24 $ \\
 \texttt{Ours} & $ 92.85 $ & $ \mathbf{93.29} $ & $ \mathbf{93.04} $ & $ \mathbf{92.66} $ & $ 86.78 $ & $ \mathbf{89.52} $ \\
 \hline
\end{tabular}
\end{table}

\begin{table}[ht!]
\caption{Experimental results for the CTC task.}
\centering
\begin{tabular}{  c | c | c | c || c | c |c }
\hline
\label{table:ctc_results}
& \multicolumn{3}{c}{ICDAR2013} & \multicolumn{3}{c}{UNLV}\\ 
\hline
Method & Prec & Recall & F1 & Prec & Recall & F1\\
\hline
\texttt{Glove} & $ 56.47 $ & $ 60.88 $ & $ 57.74 $ & $ 55.00 $ & $ 62.03 $ & $ 57.33 $ \\
 \texttt{FastText} & $ 67.85 $ & $ 64.16 $ & $ 65.46 $ & $ 63.72 $ & $ 68.85 $ & $ 65.85 $ \\
 \texttt{ConvNeXt} & $ 70.15 $ & $ 71.42 $ & $ 70.76 $ & $ 65.88 $ & $ 66.72 $ & $ 66.21 $ \\
 \texttt{ResNet50} & $ 72.81 $ & $ 69.03 $ & $ 70.16 $ & $ 72.62 $ & $ 71.83 $ & $ 71.90 $ \\
 \texttt{Ours} & $ \mathbf{87.94} $ & $ \mathbf{88.41} $ & $ \mathbf{88.17} $ & $ \mathbf{82.19} $ & $ \mathbf{85.51} $ & $ \mathbf{83.74} $ \\
 \hline
\end{tabular}
\end{table}

\subsection{Discussion and Analysis}
\label{sec:discussion_analysis}
\subsubsection{Ablation Study}
\label{sec:ablation_study}
In this section, we conduct extra experiments to demonstrate each component's effectiveness in our proposed method. Firstly, we implement two single-task models for the TSR task and CTC, respectively, using the corresponding branch design of the proposed method. Since features from the image modal, text modal, and coordinate information are utilized in the CTC task, we also implement three models that only use features from each modal. The experimental results are shown in Table~\ref{table:ablation_tsr} and Table~\ref{table:ablation_ctc}, in which $S(I), S(T), S(C), S(I+T+C)$ means the single-task model with image feature, text feature, coordinate feature, and the combined three types of features, respectively, ours means the proposed multi-task method.

The experimental results in Table~\ref{table:ablation_tsr} show that features from text modal and coordinate information are not efficient and can hardly bring any benefits when combined with features from the image modal. In contrast, visual features from the image modal are more efficient and can lead to the best performance compared with other benchmarks. Therefore, we only use visual features in the TSR task in our proposed multi-task method. Meanwhile, the experimental results in Table~\ref{table:ablation_ctc} show that visual features, text features, and coordinate information can be helpful for the CTC problem, and the combination of both types of features can contribute to a significant performance improvement. Therefore, we use visual features, text features, and coordinate information in the CTC branch of the proposed multi-task method.  

\begin{table}[ht!]
\caption{Ablation study results for the TSR task.}
\centering
\begin{tabular}{  c | c | c | c || c | c |c }
\hline
\label{table:ablation_tsr}
& \multicolumn{3}{c}{ICDAR2013} & \multicolumn{3}{c}{UNLV}\\ 
\hline
Method & Prec & Recall & F1 & Prec & Recall & F1\\
\hline
 \texttt{S(T)} & $ 36.36 $ & $ 34.56 $ & $ 34.32 $ & $ 40.03 $ & $ 35.91 $ & $ 35.94 $ \\
 \texttt{S(I)} & $ 89.42 $ & $ 91.71 $ & $ 90.52 $ & $ 84.39 $ & $ \mathbf{87.52} $ & $ 85.75 $ \\
 \texttt{S(C)} & $ 50.81 $ & $ 77.69 $ & $ 45.30 $ & $ 60.38 $ & $ 83.75 $ & $ 61.17 $ \\
 \texttt{S(T+I+C)} & $ 88.71 $ & $ 93.53 $ & $ 90.98 $ & $ 90.57 $ & $ 84.16 $ & $ 87.11 $ \\
 \texttt{Ours}& $ \mathbf{92.85} $ & $ \mathbf{93.29} $ & $ \mathbf{93.04} $ & $ \mathbf{92.66} $ & $ 86.78 $ & $ \mathbf{89.52} $ \\
 \hline
\end{tabular}
\end{table}

\begin{table}[ht!]
\caption{Ablation study results for the CTC task.}
\centering
\begin{tabular}{  c | c | c | c || c | c |c }
\hline
\label{table:ablation_ctc}
& \multicolumn{3}{c}{ICDAR2013} & \multicolumn{3}{c}{UNLV}\\ 
\hline
Method & Prec & Recall & F1 & Prec & Recall & F1\\
\hline
 \texttt{S(T)} & $ 68.11 $ & $ 62.46 $ & $ 64.48 $ & $ 72.047 $ & $ 67.85 $ & $ 69.41 $ \\
 \texttt{S(I)} & $ 70.12 $ & $ 67.52 $ & $ 68.46 $ & $ 63.98 $ & $ 76.49 $ & $ 68.62 $ \\
 \texttt{S(C)} & $ 86.07 $ & $ \mathbf{90.84} $ & $ 87.65 $ & $ 76.98 $ & $ \mathbf{85.55} $ & $ 80.74 $ \\
 \texttt{S(T+I+C)} &  $ 85.44 $ & $ 88.68 $ & $ 86.53 $ & $ 77.75 $ & $ 81.59 $ & $ 78.70 $ \\
 \texttt{Ours} & $ \mathbf{87.94} $ & $ 88.41 $ & $ \mathbf{88.17} $ & $ \mathbf{82.19} $ & $ 85.51 $ & $ \mathbf{83.74} $ \\
 \hline
\end{tabular}
\end{table}
\subsubsection{The impact of hyper parameter lambda}
\begin{figure}[htp]
\begin{center}
  \includegraphics[width=\columnwidth]{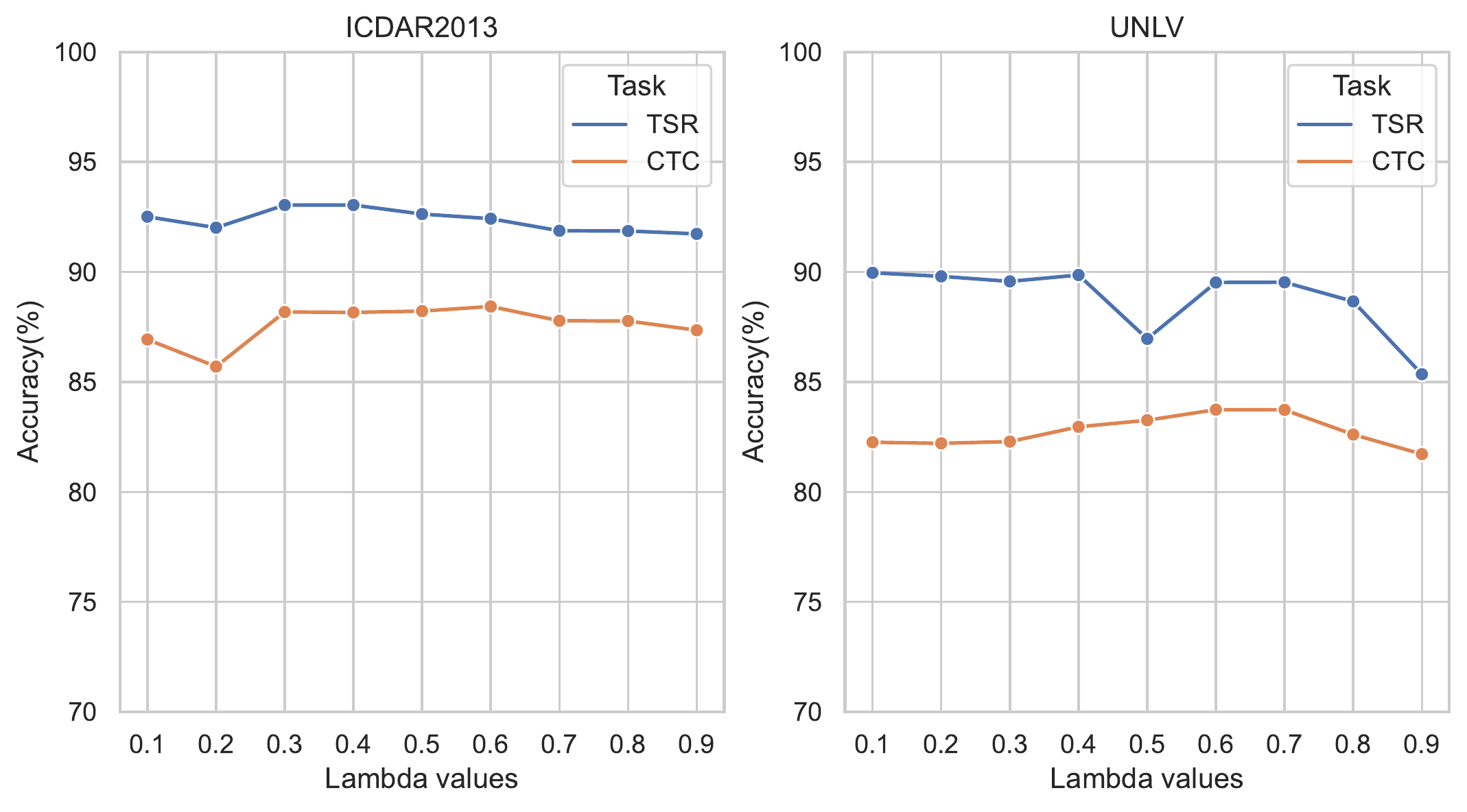}
  \caption{Experimental results with different $\lambda$ values. }
  \label{fig:lambda_performance}
\end{center}
\end{figure}

Our proposed method contains a TSR branch and a CTC branch, both of which lead to a loss, as shown in Equation~\ref{eq:loss_function}. The hyperparameter $\lambda$ is used to balance the TSR task's loss and the CTC task's loss. We conducted extra experiments to discuss the influence of this hyperparameter on the proposed model, and the results are shown in Figure~\ref{fig:lambda_performance}. The figure shows that the performance of the proposed method can be influenced by the value of $\lambda$, but can sustain an overall stable performance.  

\subsubsection{The impact of feature fusion}
As shown in Figure~\ref{fig:tsr_ctc_overall_architecture}, the features used in the CTC task come from different modalities, including text modal, image modal, and cells' coordinates, and the feature fusion layer fuse these features together before they are fed into the classifier. Therefore, we conduct extra experiments to discuss the impact of these values. The number of neurons in a fully connected layer can be any integer larger than 0. Considering the scores reported in Table~\ref{table:ablation_ctc}, and simplifying the problem, we assume $l_1 = l_2 = l_3 = l_4$ in this section. Table~\ref{table:fusion_tsr} and Table~\ref{table:fusion_ctc} present the experimental results when $l_1$ equals 16, 32, 64, 128 respectively, and the results show that increasing the number of neurons can lead to better performance to some degree, especially for TSR.

\begin{table}[ht!]
\caption{Experimental results for the the TSR task.}
\centering
\begin{tabular}{  c | c | c | c || c | c |c }
\hline
\label{table:fusion_tsr}
& \multicolumn{3}{c}{ICDAR2013} & \multicolumn{3}{c}{UNLV}\\ 
\hline
Method & Prec & Recall & F1 & Prec & Recall & F1\\
\hline
  $l_{16}$ & $ 81.24 $ & $ 92.57 $ & $ 85.86 $ & $ 81.65 $ & $ 82.25 $ & $ 81.65 $ \\
  $l_{32}$ & $ 84.75 $ & $ \mathbf{93.12} $ & $ 88.49 $ & $ 79.77 $ & $ 82.86 $ & $ 80.55 $ \\
  $l_{64}$ & $ \mathbf{85.69} $ & $ 92.51 $ & $ \mathbf{88.75} $ & $ 79.85 $ & $ \mathbf{85.74} $ & $ 82.53 $ \\
  $l_{128}$ & $ 80.13 $ & $ 93.60 $ & $ 85.54 $ & $ \mathbf{87.41} $ & $ 85.66 $ & $ \mathbf{86.45} $ \\
 \hline
\end{tabular}
\end{table}

\begin{table}[ht!]
\caption{Experimental results for the CTC task.}
\centering
\begin{tabular}{  c | c | c | c || c | c |c }
\hline
\label{table:fusion_ctc}
& \multicolumn{3}{c}{ICDAR2013} & \multicolumn{3}{c}{UNLV}\\ 
\hline
Method & Prec & Recall & F1 & Prec & Recall & F1\\
\hline
$l_{16}$ & $ 79.76 $ & $ 82.48 $ & $ 78.83 $ & $ \mathbf{82.84} $ & $ 81.94 $ & $ 82.26 $ \\
  $l_{32}$ & $ 83.54 $ & $ 83.50 $ & $ 82.29 $ & $ 81.02 $ & $ 85.04 $ & $ 82.92 $ \\
  $l_{64}$ & $ 89.47 $ & $ 81.22 $ & $ 84.72 $ & $ 80.76 $ & $ 87.09 $ & $ \mathbf{83.57} $ \\
  $l_{128}$ & $ \mathbf{90.83} $ & $ \mathbf{84.33} $ & $ \mathbf{87.22} $ & $ 79.36 $ & $ \mathbf{87.18} $ & $ 82.86 $ \\
 \hline
\end{tabular}
\end{table}

\section{Conclusion and Future Work}
\label{sec:conclusion}
Participants in the ICT supply chains generate "big tabular data" that is difficult to manage but often contains valuable information.  This paper introduced a multi-task model that simultaneously solves the TSR and CTC problems, which are critical steps in transforming tabular data into a machine-interpretable format. Experimental results show that both TSR and CTC tasks can benefit from the multi-task design compared to their single-task counterparts. Besides, the proposed method can outperform the state-of-the-art benchmark models by a large margin, increasing the F1 score from 70.76\% to 88.17\%, and from 71.90 to 83.74 for the CTC task on the ICDAR2013 and UNLV dataset respectively. For the TSR task, increase the F1 score from 92.30\% to 93.04\% on the ICDAR2013 dataset and from 87.24\% to 89.51\% on the UNLV dataset. We tackle the TSR and CTC simultaneously with a single multi-task model, whereas the two problems can also be addressed in sequence in future work.

\bibliographystyle{IEEEtran}

\end{document}